\documentclass[10pt,twocolumn,letterpaper]{article}

\usepackage{iccv}
\usepackage{times}
\usepackage{epsfig}
\usepackage{graphicx}
\usepackage{amsmath}
\usepackage{amssymb}
\usepackage{multirow}
\usepackage{pifont}
\usepackage[table,xcdraw]{xcolor}
\usepackage[accsupp]{axessibility}  
\usepackage{authblk}

\newtheorem{definition}{Definition}

\usepackage[pagebackref=true,breaklinks=true,letterpaper=true,colorlinks,bookmarks=false]{hyperref}

 \iccvfinalcopy 

\def\iccvPaperID{2367} 

\ificcvfinal\fi

\newcommand{\ours}{Chamfer OOD examples}

\newcommand{\oursshort}{CODEs}
\newcommand{\oursshortSingle}{CODE}
\newcommand{\ourssmall}{CODEs}
\newcommand{\FP}{OOD examples}

\newcommand{\ourscommon}{OOD examples}

\newcommand{\TODO}[1]{{\color{red}{[TODO: #1]}}}
\newcommand{\tang}[1]{{\color{red}{#1}}}

\newcommand{\peng}[1]{{\color{red}{[Weilong: #1]}}}

\newcommand{\go}[1]{{\color{blue}{[Gordon: #1]}}}
\newcommand{\revised}[1]{{\color{magenta}{#1}}}
\newcommand{\firstpara}[1]{\noindent\textbf{{#1}.}~~}

\newcommand{\pozhehao}{\kern0.3ex\rule[0.8ex]{2em}{0.1ex}\kern0.3ex}
\usepackage{enumitem}
\usepackage{booktabs}
\usepackage{footnote}

\usepackage{makecell}

\newcommand{\ourline}[1]{\toprule[#1]}

\begin{document}

\title{CODEs: Chamfer Out-of-Distribution Examples against  Overconfidence Issue}

\makeatletter 
\renewcommand\AB@affilsepx{\quad \protect\Affilfont} 
\makeatother
\author[1]{Keke Tang\thanks{Joint first authors.}}
\author[1,2]{Dingruibo Miao\textsuperscript{$\ast$}}
\author[1]{Weilong Peng\thanks{Corresponding authors: wlpeng@gzhu.edu.cn,  zqgu@gzhu.edu.cn.}}
\author[1]{Jianpeng Wu}
\author[1]{Yawen Shi}
\author[1]{Zhaoquan Gu\textsuperscript{\textdagger}}
\author[1]{\\Zhihong Tian}
\author[3,4]{Wenping Wang}
\affil[1]{Guangzhou University}
\makeatletter 
\renewcommand\AB@affilsepx{\\ \protect\Affilfont} 
\makeatother
\affil[2]{Peng Cheng Laboratory}
\makeatletter 
\renewcommand\AB@affilsepx{\quad \protect\Affilfont} 
\makeatother
\affil[3]{Texas A\&M University}
\makeatletter 
\renewcommand\AB@affilsepx{\quad \protect\Affilfont} 
\makeatother
\affil[4]{The University of Hong Kong}
\makeatletter 
\renewcommand\AB@affilsepx{\quad \protect\Affilfont} 
\makeatother
\renewcommand\Authands{, }

\maketitle
\ificcvfinal\thispagestyle{empty}\fi

\begin{abstract}

Overconfident predictions on out-of-distribution (OOD) samples is a thorny issue for deep neural networks.
The key to resolve the OOD overconfidence issue inherently is to build a subset of  OOD    samples  and  then  suppress  predictions  on them.
This paper proposes
the 
\ours{} (\oursshort{}),  whose distribution is close to that of in-distribution samples, and thus could be utilized  to alleviate the OOD overconfidence issue effectively by suppressing predictions on them.
To obtain \oursshort{}, 
we  first generate
seed OOD examples
via slicing\&splicing operations on in-distribution samples from  different categories, and then  feed them to the Chamfer generative adversarial network for distribution transformation,
without  accessing to any extra data. 
Training with suppressing predictions on  \oursshort{}  is validated to alleviate the  OOD overconfidence issue largely without hurting  classification accuracy, and  outperform the state-of-the-art methods.
Besides, we  demonstrate   \oursshort{} are  useful for improving OOD detection and  classification. 

\end{abstract} 
\section{Introduction}


Deep neural networks (DNNs) have obtained state-of-the-art performance in the classification problem~\cite{He-2015-SurpassClassification}.
Since  those classification systems are generally designed for a static and closed world~\cite{Bendale-2015-OpenSet},  DNN classifiers will  attempt to make predictions even with the occurrence of new concepts in real world.
Unfortunately, those  unexpected predictions are likely to be overconfident.
Indeed, a growing body of evidences show that  DNN classifiers suffer from the OOD overconfidence issue of
being fooled easily to generate
overconfident predictions on OOD samples~\cite{nguyen-2015-EasilyFooled,Goodfellow-2015-AdversarialExamples}.

A widely adopted solution is to calibrate the outputs between in- and out-of-distribution samples to make them easy to detect~\cite{Hendrycks-2017-Detecting-Out-of-Distribution,Liang-2018-EnhancingOut-of-distribution}.
By this way, overconfident predictions on OOD samples could be rejected as long as identified by  OOD detectors.
While those approaches are significant steps towards reliable classification, 
the OOD overconfidence issue of DNN classifiers  remains unsolved.
Besides, as challenged by Lee et al.~\cite{Lee-2018-CalibrationOutOfDistribution}, the performance of OOD detection highly depends on DNN classifiers and they fail to work if the classifiers do not separate the  predictive distribution well.
This motivates us to resolve the OOD overconfidence issue inherently with enforcing DNN classifiers to make low-confident predictions on OOD samples.


\begin{figure}
\centering
\vspace{-2mm}
  \includegraphics[width=0.99\linewidth]{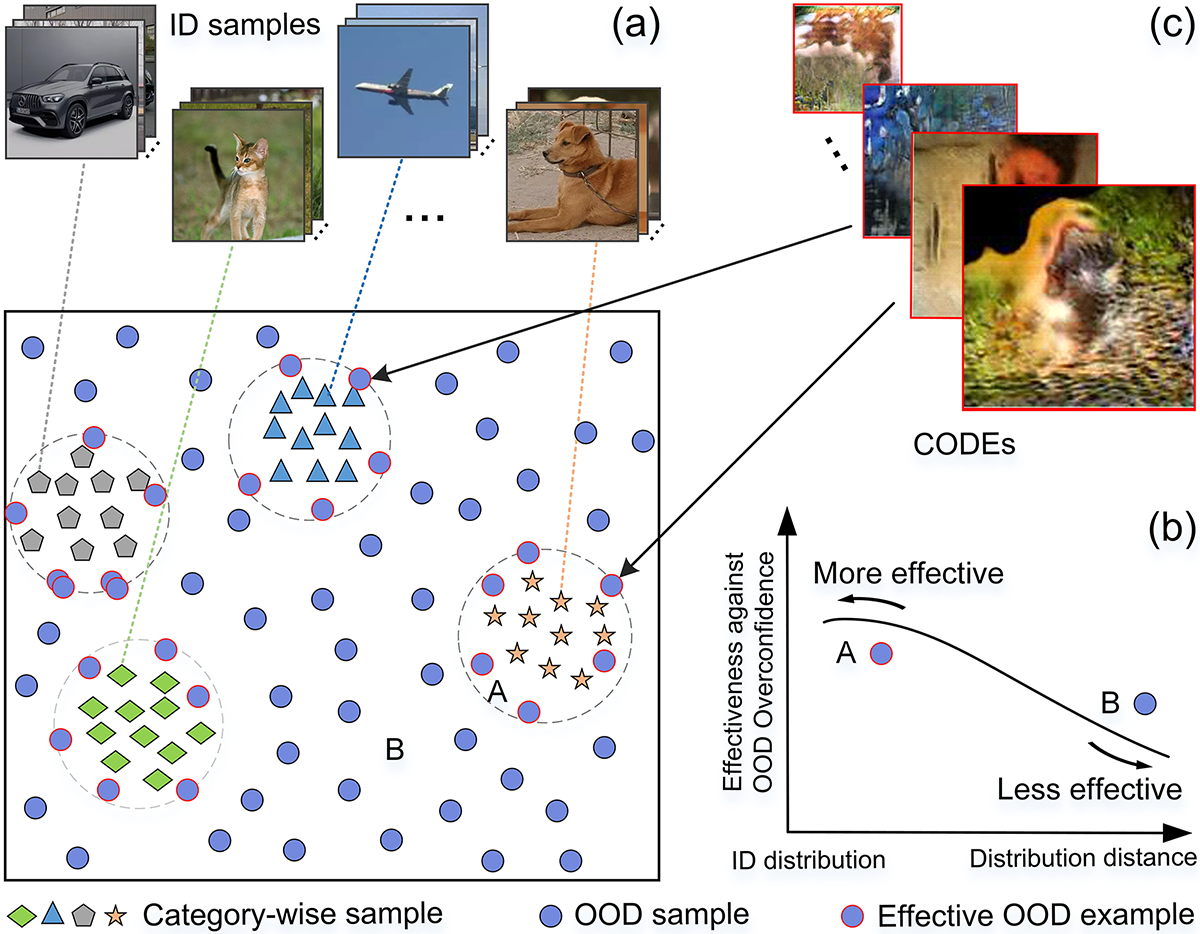}
  \caption{
For a classification task,
  (a)  OOD samples can be infinitely outnumber  ID  samples;
  (b)  intuitively,  OOD samples, whose distribution  is closer to that of  ID  samples, are more  likely to be effective  against the OOD overconfidence issue;
  (c) we aim at generating \oursshort{},  a kind of effective \FP{}.
  }
  \vspace{-3mm}
  \label{fig:teaser}
\end{figure}

Since with  infinite amount, the key to resolve the OOD overconfidence issue is to build a subset of  OOD
samples and then suppress predictions on  them. 
Lee et al.~\cite{Lee-2018-CalibrationOutOfDistribution} use a generative adversarial network~\cite{goodfellow2014generative} to model the subset which, however, requires to be tuned on the testing-distribution.
Without synthesizing data from a carefully designed distribution,  Hendrycks et al.~\cite{Hendrycks-2018-AnomalyDetection} adopt an auxiliary dataset to simulate the subset. 
However, the optimal choice of such dataset remains an open question, challenges of data imbalance and computational complexity make it less efficient and practical~\cite{Li-2020-backgroundResample}.
In contrast, Hein et al.~\cite{Hein-2019-WhyOutOfDistribution}  simply adopt  random noises  and permuted images, and report promising results.
It thus brings us to the main topic of this paper (see Fig.~\ref{fig:teaser}): 
can we get a subset of OOD samples that is more effective for alleviating the OOD overconfidence issue
by suppressing predictions on them? 
Intuitively, suppressing a subset of OOD samples whose distribution is close to that of in-distribution (ID) samples, are expected to bring more benefits, 
since they are harder to be differentiated by DNNs to make low-confident predictions. 
Under this hypothesis, we  propose to generate a subset of  OOD samples whose distribution is close  to that of ID samples, i.e., effective OOD examples.

In this paper, we propose  the novel \ours{} (\oursshort{}),  which is a kind of effective OOD examples.
Besides, we devise a simple yet effective method to generate \oursshort{} with training data only.
Specifically, we first generate seed examples that is OOD  by slicing\&splicing operations,
and then feed them into the Chamfer generative adversarial network (Chamfer GAN) for distribution transformation.
Particularly, the Chamfer distance loss  is intentionally imposed on Chamfer GAN to maintain the pixel-level statistic of seed examples, such that \oursshort{} remain to be OOD.
We   validate the effectiveness of our approach by suppressing predictions on  \oursshort{} during training.
 Extensive experiments show that the OOD overconfidence issue will be largely alleviated by our approach without hurting the original classification accuracy on ID samples,
 and that our approach outperforms the  state-of-the-art methods.
We also demonstrate \oursshort{} could have broad applications, e.g.,  to improve  OOD detectors and image classification.

Overall, our contribution is  summarized as follows:

\begin{itemize}

\setlength{\itemsep}{0pt}
\setlength{\parsep}{2pt}
\setlength{\parskip}{0pt}

\item
We  show distribution distance is the key factor for OOD examples in alleviating the OOD overconfidence issue, with  many other factors excluded.

\item
We propose a simple yet effective method  
based on slicing\&splicing operations and  Chamfer GAN to generate \oursshort{} without accessing to any extra data.

\item
We validate the superiority of \oursshort{} in alleviating the OOD overconfidence issue of DNN classifiers  inherently without hurting the classification accuracy.

\item
We demonstrate the effectiveness of \oursshort{}  in  improving OOD detection and image classification. 


\end{itemize}

\if
including security-critical ones, such as autonomous cars~\cite{Li-2019-AutoCar}, face identification~\cite{Ranjan-2019-FaceIdentification} and robot task planing~\cite{Paxton-2019-visualPlanning},
making the security aspect of DNNs increasingly important.
\fi

\section{Related Work}

\noindent{\bf{Suppressing Predictions on OOD Samples.}} 
To suppress predictions on OOD samples, Lee et al.~\cite{Lee-2018-CalibrationOutOfDistribution}  trained a classifier as well as a GAN that models the boundary of in-distribution samples, and enforced the classifier to have lower confidences on GAN samples.
However, for each testing distribution, they tuned the classifier and GAN using samples from that out-distribution.
Without  accessing to the testing distribution directly, Hendrycks et al.~\cite{Hendrycks-2018-AnomalyDetection}  used an auxiliary dataset disjoint from test-time data to simulate it.
Meinke et al.~\cite{Meinke-2020-CCU} explicitly integrated a generative model and provably showed that the resulting neural network produces close to uniform predictions far away from the training data. 
However, for suppressing the predictions, they also adopted auxiliary datasets.
Since challenges of data imbalance  and computational complexity brought by auxiliary datasets~\cite{Li-2020-backgroundResample}, Hein et al.~\cite{Hein-2019-WhyOutOfDistribution} proposed to simply 
consider random noise and permuted images as OOD samples.

We also aim to generate OOD samples with training data only. 
Differently, we intentionally generate effective  OOD examples, and thus  obtain better results.
Besides, our approach would not hurt classification accuracy on ID samples, which is not guaranteed by using auxiliary datasets.

\noindent{\bf{OOD Detection.}} 
Hendrycks et al.~\cite{Hendrycks-2017-Detecting-Out-of-Distribution} built the first  benchmark for OOD detection and evaluated the simple threshold-based detector.
Recent  works 
improve OOD detection by using  the ODIN score~\cite{Liang-2018-EnhancingOut-of-distribution,Hsu-2020-generalODIN},  Mahalanobis distance~\cite{Lee-2018-Mahalanobis},  
energy score~\cite{Liu-2020-energyOOD},  ensemble of multiple classifiers~\cite{vyas2018-ensembleClassifiers,Yu-2019-MaximumClassifierDiscrepancy}, residual flow~\cite{Zisselman-2020-residualFlow}, generative
models~\cite{ren2019likelihood}, self-supervised learning~\cite{Hendrycks-2019-selfsupervised,Mohseni-2020-selfSuperviseOOD} and gram matrices~\cite{Sastry-2020-OODGramMatric}.

All above approaches detect whether a test sample is from in-distribution (i.e., training distribution by a classifier) or OOD, and is usually deployed  with  combining the original $n$-category classifier to handle recognition in the real world~\cite{Bendale-2016-OpenSetRecognition,Yoshihashi-2019-OpenSetRecognition}.
Differently, our  motivation is to enforce the original $n$-category classifier to make low-confident predictions on OOD samples inherently.


\noindent{\bf{Confidence Calibration.}}
The calibration of the confidence of predictions are traditionally considered on the true input distribution.
Gal and Ghahramani~\cite{Gal-2016-dropoutCali} adopted Monte Carlo dropout to estimate the single best uncertainty by interleaving  DNNs with Bayesian models.
Lakshminarayanan et al~\cite{lakshminarayanan2017simple} used ensembles of networks to obtain combined uncertainty estimates.
Guo et al.~\cite{Guo-2017-OnCalibration}  
utilized temperature scaling to obtain calibrated probabilities.

Differently, our work focus on  calibrating the confidence of predictions on OOD samples.
Besides, it has been validated that models for confidence calibration on the input distribution cannot be used for out-of-distribution~\cite{leibig2017leveraging}.

\noindent{\bf{Data Augmentation.}}
Data augmentation is originally designed to prevent networks from overfitting by synthesizing label-preserving images
~\cite{krizhevsky2012imagenet,devries-2017-cutout}.
Another type
is to improve  adversarial robustness or classification performance by adding adversarial noise to the training images~\cite{madry-2018-PGDAttack,Xie-2020-adversarial-improve-recognition}.

Differently, we suppress the predictions on the augmented images which are OOD, while 
augmented ID images  in their methods are trained in the same manner as original training data.

\section{Effective OOD Examples}
\label{sec:problem}

\firstpara{Preliminary}
This work considers the setting in multi-category classification problems. Let  $\mathcal{A}$ be the set of all digital images under consideration and $\mathcal{I}\subseteq \mathcal{A}$ be the set of all in-distribution samples that could be assigned a label in $\{1,2,...,K\}$.
Then, $\mathcal{A}\setminus\mathcal{I}$ is the set of  all OOD samples.
Specifically, we have a classifier $f: \mathcal{I} \rightarrow \{1,2,...,K\}$ that could give a prediction for any image in $\mathcal{A}$.

Indeed, the classifier $f$ could make  high-confident predictions on images in $\mathcal{A}\setminus\mathcal{I}$. 
It is thorny when DNN classifiers face an open-set world.
Since OOD samples can be infinitely many, suppressing them all is impractical. 
We thus aim at collecting a subset of OOD samples that are effective for alleviating the OOD overconfidence issue by suppressing predictions on them, i.e., effective OOD examples.

\begin{definition}
\label{def:x}
\noindent
{\bf{Effective OOD Examples}}. 
Given a small constant $\delta$, an effective OOD example $x$ is any image in $\mathcal{S}_{eo}$ satisfying 
\[\mathcal{S}_{eo} \triangleq  \{ x \in \mathcal{A}\setminus\mathcal{I} \ | \  D(P_\mathcal{I},P_{eo}) \le \delta \},\]
where $P_\mathcal{I}$ is the distribution of  $\mathcal{I}$,
$P_{eo}$ is the distribution of  $\mathcal{S}_{eo}$,
and $D$ is the criterion to measure the distance between two distributions.
\end{definition}


Since $P_{eo}$ is close to that of in-distribution  samples $P_\mathcal{I}$, it is hard to differentiate them.
Therefore,  suppressing effective OOD examples is expected to bring more benefits than suppressing the others that are easier to be differentiated. 

\section{Method}
\label{sec:method}

In this section, we will introduce the approach to obtain effective OOD examples with training data only.
Particularly, we start by generating seed  examples that are OOD,
 then  convert those seeds into \ours{} (\oursshort{}) by enforcing the distribution restriction.
Finally,  we will demonstrate  how to  alleviate the OOD overconfidence issue by
training with \oursshort{}.

\subsection{Generating Seed Examples}

We generate seed examples by splicing  local patches from images with multiple different categories via two key operations.



\begin{figure}[!t]
\centering
  \includegraphics[width=0.99\linewidth]{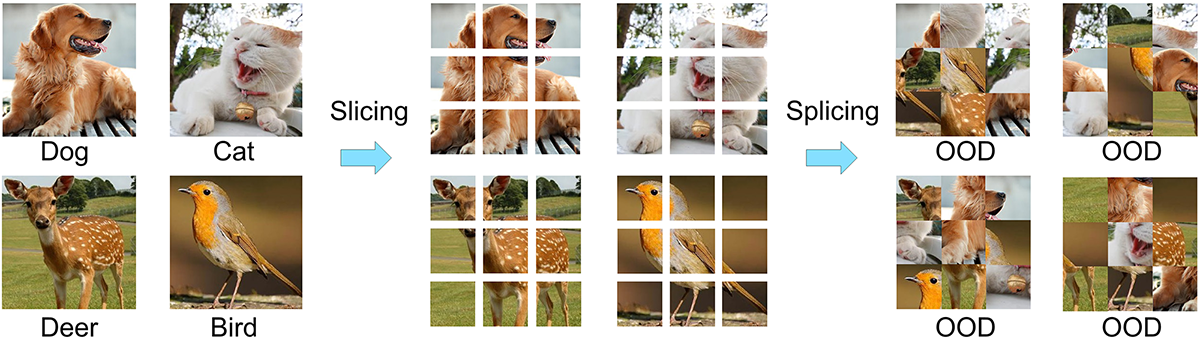}
  \caption{Generate seed examples by slicing and splicing ($k=3$).}
  \vspace{-4mm}
  \label{fig:cropAndPaste}
\end{figure}

{\bf{Slicing Operation}}:
\begin{equation}
\begin{aligned}
    \{p_0^{x}, p_1^{x}, ..., p_{k*k-1}^{x}\} = & OP_{slice}(x, k) \\
\end{aligned}
\end{equation}
where $x$ is an image in the training set and   $p_{t}^{x}$  is  the $t$-th piece of $x$.
By this operation, each image is divided into $k\times k$ numbers of  patches  equally (see Fig.~\ref{fig:cropAndPaste}).     

{\bf{Splicing Operation}}:
\begin{equation}
\begin{aligned}
& \overline{x}  = OP_{splice}([p_0^{x^0}, p_1^{x^1}, ..., p_{k*k-1}^{x^{k*k-1}}], k) \\  
&   st. \  \ ( c(x^0)==c(x^1)==...==c(x^{k*k-1})) = False
\end{aligned}
\end{equation}
where $k\times k$ numbers of  patches with uniform size  that are sliced from images with different categories (denoted by $c(\cdot)$)
are spliced into an image with the same size as $x$ (see Fig.~\ref{fig:cropAndPaste}). 
Particularly, those patches are required to be not from  images with the same categories.

Since the resulting seed examples do not belong to any categories visually  (see Fig.~\ref{fig:cropAndPaste}), and thus are OOD examples.
Note that, we just provide two simple operations for the splicing of image patches.
More complex operations (e.g., considering angles, scales and rotation)  could be extended easily, but is out of the scope of this paper.

\begin{figure*}[t!]
\centering
  \vspace{-1mm}
  \includegraphics[width=0.98\linewidth]{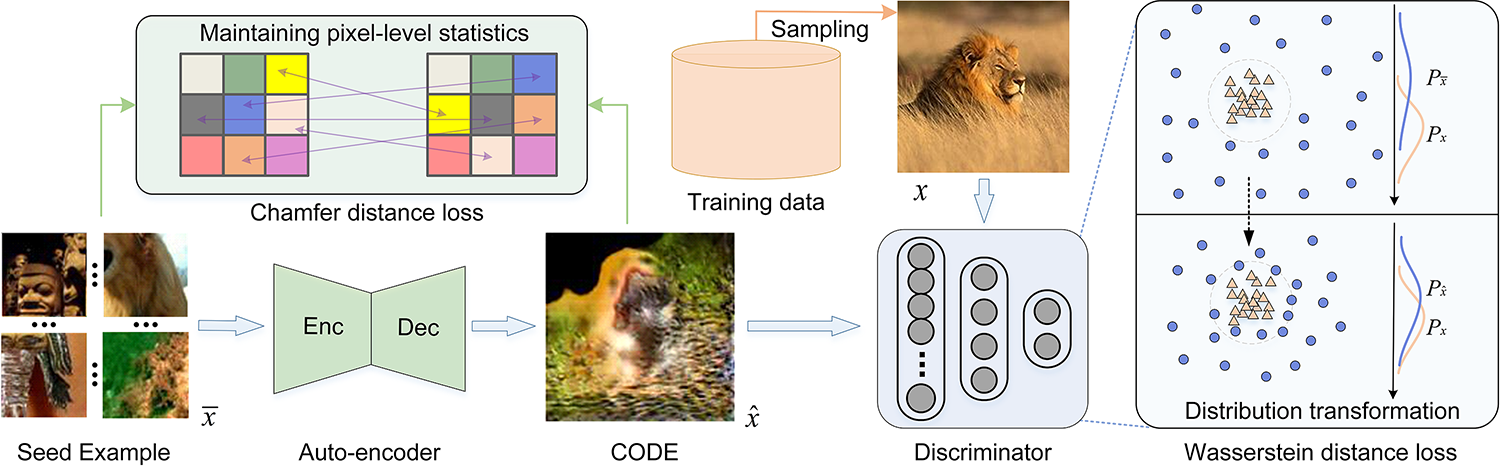}
  \caption{
  The framework of Chamfer GAN:  given a seed example $\bar{x}$ as input, the auto-encoder backbone outputs the CODE $\hat{x}$; it is supervised by the Chamfer distance loss for maintaining pixel-level statistics and the  Wasserstein distance loss for  transforming the distribution $P_{\hat{x}}$ to  $P_{{x}}$.
}
  \vspace{-1mm}
  \label{fig:ae_gan}
\end{figure*}

\subsection{Generating \ourssmall{}}

 We  transform the distribution of seed examples to that of the training data by feeding them into a novel Chamfer generative adversarial network (Chamfer GAN) with maintaining pixel-level statistics, to generate \oursshort{}. 
In this following, we will describe the architecture design of Chamfer GAN first, and then discuss why \oursshort{} generated by Chamfer GAN  are effective OOD examples.


\subsubsection{Chamfer GAN}

For the design of Chamfer GAN, we adopt the popular auto-encoder~\cite{Hinton2006-reducing} as our backbone, see Fig.~\ref{fig:ae_gan}. Given a seed example $\bar{x}$ as input, the encoder $Enc$ projects it into high-dimensional compact space, and the decoder $Dec$ decode the projected feature  to  reconstruct the  \oursshortSingle{} $\hat{x}$. 
Particularly, we adopt the Wasserstein distance loss  in WGAN~\cite{Gulrajani-2017-WGAN} to {\em transform the data distribution}, and the Chamfer distance loss ~\cite{Borgefors-1988-chamfer} to  {\em maintain pixel-level statistics}.

\firstpara{Wasserstein distance loss}
To enforce the distribution of $\hat{x}$ (i.e., $P_{\hat{x}}$) to be close to that of the training data $x$ (i.e., $P_{x}$), we adopt the same adversarial loss as in WGAN~\cite{Gulrajani-2017-WGAN}
with the gradient penalty term omitted for clarity, which is defined as:

\begin{equation}
\label{eq:wgan}
L_\text{WD} = {\mathbb{E}_{x \sim P_{x}}}  [Dis(x)] - {\mathbb{E}_{\hat{x} \sim P_{\hat{x}}}} [Dis(\hat{x})]   
\end{equation}

\firstpara{Chamfer distance loss} 
To facilitate maintaining  pixel-level statistics during the reconstruction process, we adopt the Chamfer distance  for restriction, which is defined as:

\begin{small}
\begin{equation}\label{eq:general_cd}
L_{\text{CD}} =  {{\mathbb{E}}_{\bar{x}{\sim}P_{\bar{x}}}}  \bigg({ \sum_{\hat{x_i} \in \hat{{x}}} \min_{\bar{x_j} \in \bar{x}} \|\hat{x_i}-\bar{x_j}\|_2^{2}}  + \sum_{\bar{x_j} \in \bar{x}} \min_{\hat{x_i} \in \hat{{x}}} {\|\hat{x_i}-\bar{x_j}\|_2^2}\bigg) 
\end{equation}
\end{small}
\begin{equation}\label{eq:general_cd_cons}
  \hat{x} = Dec(Enc(\bar{x}))
\end{equation}
where $P_{\bar{x}}$ is the distribution that seed example  $\bar{x}$ is in, $\bar{x_j}$ and $\hat{x_i}$ denote the pixel in $\bar{x}$ and $\hat{x}$ respectively. 
Note that, we do not require $\bar{x}$  and $\hat{x}$  to be  extremely the same which is enforced by $L_2$ loss in traditional auto-encoder.
Instead, the Chamfer distance loss enforces each pixel in   $\bar{x}$   to have a corresponding pixel in $\hat{x}$ outputted by $Enc$-$Dec$, 
but could  at a different location, namely with pixels rearranged. Therefore, the  pixel-level statistics is maintained.

By combining the above two loss functions, Chamfer GAN could transform the distribution of ${\hat{x}}$ to  be close to that of the training data $x$, while maintaining the pixel-level statistics of $\bar{x}$.
The final loss function is thus as follows:
\begin{equation}\label{eq:chamferGAN}
L_{\bar{x}\rightarrow\hat{x}}  = {\alpha}L_{\text{WD}} +  L_{\text{CD}}
\end{equation}
\noindent{}where $\alpha$ is a scalar weight set by $1.0\mathrm{e}{-5}$ by default. 
For the training of Chamfer GAN, we train the classifier $Dis$ and $Enc$-$Dec$ in an iterative manner as in WGAN.
Note that, the Chamfer distance loss is critical for Chamfer GAN.
Specifically, Chamfer distance loss restricts the distribution transforms along a special feature space whose corresponding image space is a ``pixel rearranged space''. 

By feeding seed examples into Chamfer GAN, 
\oursshort{} are obtained that  maintain  pixel-level statistics of  seed examples, but within a distribution  much closer to that of the training data.


\subsubsection{Discussion on \oursshort{}}

Since with  supervision by the Wasserstein distance loss, the  distribution of \oursshort{} is transformed to be close to that of the training data.
Besides, \oursshort{} remain to be  OOD, due to:  1)  the pixel-level statistics of  seed examples that are originally OOD are maintained by the Chamfer distance loss; 
and 
2) the training of WGAN that transforms the distribution to be the same as that of the training data is hard originally, and it is even harder with the restriction of Chamfer distance loss.
Overall,  \oursshort{} 
are effective \ourscommon{}.
Please refer to Sec.~\ref{subsec:CODE} for validation.

\subsection{Using \oursshort{} against OOD Overconfidence}

\oursshort{} could be utilized to alleviate the OOD overconfidence issue by suppressing predictions on them over each category, namely enforcing  averaged confidences over all categories (i.e., $\frac{1}{K}$) with the  following loss function:

\begin{equation}\label{adv_train}
  L_{sup}(\hat{x}) =    
   \sum_{i=1}^{i=K} \frac{1}{K} \log{V_i(\hat{x})}
\end{equation}

\noindent
where  $K$ is the category number and $V_i(\hat{x})$ is the normalized prediction confidence of $\hat{x}$ over category $i$.

For training, we adopt 50\% images from the original training set supervised with the cross-entropy loss, while the others are \oursshort{} supervised with Eqn.~\ref{adv_train}.

\section{Experiments}
\label{sec:Experiments}

\begin{figure*}[t!]
\centering
  \includegraphics[width=0.97\linewidth]{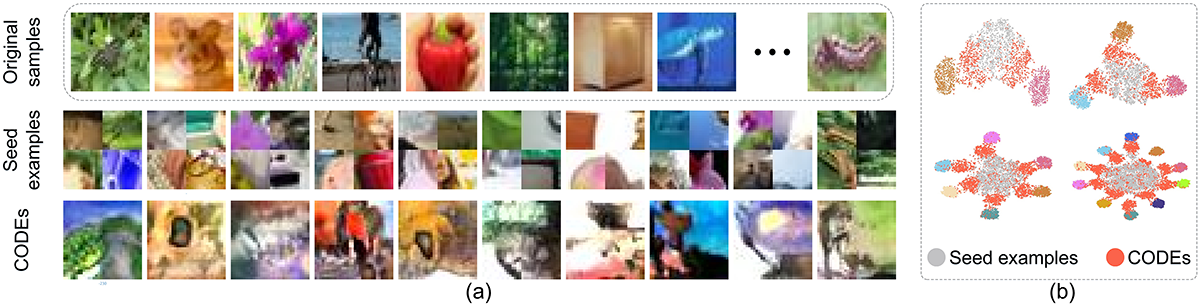}
  \vspace{-1mm}
  \caption{(a) Row-1: original images from CIFAR-100; Row-2: seed examples; Row-3:  the corresponding \ourssmall{} of Row-2.  (b) The t-SNE embeddings of original images  (2, 3, 6 and 10 categories), seed examples and \oursshort{} on CIFAR-10.}
  \vspace{-1.5mm}
  \label{fig:vis_adv}
\end{figure*}

This section includes four parts. 
Firstly, we  analyze the features of \oursshort{}. 
Secondly, we   extensively evaluate \oursshort{} for alleviating the OOD overconfidence issue inherently
with  comparing with the state-of-the-art methods.
Thirdly, we  demonstrate the  applications of \oursshort{}, e.g., for improving OOD detection and classification.
Finally, we  report the ablation studies. 

\subsection{Features of \oursshort{}}
\label{subsec:CODE}
\firstpara{Implementation} 
We set $k$ as 2 for generating seed examples. 
The auto-encoders in Chamfer GAN for 32$\times$32 and 28$\times$28 images adopt four convolution layers  to project images to the resolution  of 2$\times$2   with channel number of 512,
while the decoder  is symmetric to the encoder, except replacing the convolutions  with transposed ones.
For $224\times224$ images, we adopt the architecture in~\cite{Badrinarayanan-2017-segnet}.
We train the Chamfer GAN with a batch size of 32 for 1800 epochs on CIFAR-10 and CIFAR-100, 800 epochs on SVHN, MNIST, FMNIST and CINIC-10, and 50 epochs on  ImageNet. The optimizer and learning rate are the same as in WGAN~\cite{Gulrajani-2017-WGAN}. For more details, please refer to the supplementary.

\firstpara{Visualization} 
Fig.~\ref{fig:vis_adv}(a) visualizes  seed examples, the corresponding \oursshort{}, 
and original images from CIFAR-100.
Visually, \oursshort{} are more natural than seed examples, but we could not induce any categories on them, indicating that they are OOD examples. 
Fig.~\ref{fig:vis_adv}(b) visualizes the t-SNE~\cite{2008Visualizing} embeddings of original images, seed examples and \oursshort{} on CIFAR-10.
Specifically, the embeddings are based on the features outputted by the last convolutional layer of ResNet-18. 
It demonstrates that the distributions of \oursshort{} are much closer to that of   clusters of original images compared with seed examples, validating the usefulness of Chamfer GAN for distribution transformation.

\firstpara{Distribution Distance and Prediction Confidence}
We report the distribution distances between seed examples, \oursshort{} and the original images in CIFAR-10, CIFAR-100 and SVHN, measured by Fréchet Inception Distance (FID)~\cite{Heusel-2017-FID} in Tab.~\ref{tab:chamfer}.
It could be seen that the distribution distances between \oursshort{} and original images are much smaller than that of seed examples, validating the process  of distribution transformation. 
Particularly, with a closer distribution to that of the original images, 
\oursshort{} are  predicted  by ResNet-18 with higher confidences.

\begin{table}[t!]
\centering
\scalebox{0.83}{ 
\begin{tabular}{cc|ccc}
\hline
\ourline{1pt}

                     &                & CIFAR-10 & CIFAR-100 & SVHN  \\ 
                     \hline
\multirow{2}{*}{FID } & Seeds examples & 50.49    & 48.83     & 83.51 \\
                     & CODEs          & 36.53    & 36.53     & 77.94 \\ \hline
\multirow{3}{*}{MMC (\%)} & Origin         & 98.40     & 91.28      & 99.42  \\
                     & Seeds examples & 72.53     & 38.90      & 72.37  \\
                     & CODEs          & 74.84     & 40.20       & 78.74  \\ 
\ourline{1pt}
\end{tabular}
}
\caption{ Distribution distances of seed examples and \oursshort{} to that of original datasets measured by FID, and the prediction confidences on them measured by mean maximum confidence (MMC).
}
\label{tab:chamfer}
\end{table}

\subsection{Alleviating OOD Overconfidence by \oursshort{}}

\begin{table*}[]
\centering
 \scalebox{0.9}{
\begin{tabular}{c|c|l|cccccccc}
\ourline{1pt}
{train set} &
  \multicolumn{1}{c|}{{metric}} &
  {test set} &
  {Baseline} &
  {CEDA} &
  {ACET} &
  {CCUs} &
  {Ours} &
  {Ours++} &
  {OE} &
  {CCUd}\\ \hline

\multicolumn{3}{c|}{{with auxiliary dataset}} &
  \multicolumn{1}{l}{} &
  \multicolumn{1}{l}{} &
  \multicolumn{1}{l}{} &
  \multicolumn{1}{l}{} &
  \multicolumn{1}{l}{} &
  \multicolumn{1}{l}{} &
   \multicolumn{1}{c}{ \checkmark} &
  \multicolumn{1}{c}{\checkmark} 

\\ \hline

\multirow{9}{*}{{\rotatebox{90}{CIFAR-10}}} &
  TE &
  CIFAR-10 &
\textbf{5.38} &
\textbf{5.38 }&
  5.61 &
  5.56 &
  5.58 &
  5.52 &
 5.75 &
  6.01 \\ \cline{2-11}
 &
  ID MMC &
 
  CIFAR-10  &
  97.04 &
 \textbf{ 97.51} &
 96.60 &
  97.27 &
  97.15 &
  96.77 &
   88.87 &
  80.26 
  \\\cline{2-11}
 &
  \multirow{7}{*}{OOD MMC } 
  
   &
  SVHN &
  73.38 &
  72.46 &
  70.69 &
  74.13 &
  62.37 &
  47.18 &
  10.48&
  \underline{ 10.41} \\
 &
   &
  CIFAR-100 &
  79.47 &
  80.54 &
  79.28 &
  80.83 &
  70.14 &
 59.52&
  21.62&
 \underline{ 16.98}  \\
 &
   &
  LSUN\_CR &
  73.38 &
  75.15 &
  75.89 &
  75.95 &
  64.11 &
  53.67 &
  10.56 &
  \underline{ 10.38} 
  \\
 &
   &
  Noise &
  69.24 &
  10.36 &
  10.62 &
  77.85 &
  \textbf{10.27} &
  10.89 &
   13.37&
   10.36
  \\
 &
   &
  Uniform &
  99.49 &
  73.23 &
  \textbf{10.00} &
  \textbf{10.00} &
  65.30 &
  \textbf{10.00} &
  10.35&
  \underline{ 10.00} 
   \\
 &
   &
  Adv. Noise &
  100.00 &
  98.50 &
  11.20 &
  10.01 &
  15.80 &
  10.29 &
  100.00 &
 \underline{ 10.00} 
  \\
 &
   &
  Adv. Samples &
  100.00 &
  100.00 &
  63.30 &
  - &
  54.40 &
  \textbf{27.79} &
  - &
  - \\ \hline
\multirow{9}{*}{{\rotatebox{90}{CIFAR-100}}} &
  TE &
  CIFAR-100 &
  23.34 &
  23.54 &
  24.01 &
  24.13 &
   \textbf{23.22} &
  23.55 &
  25.51&
  26.53 
   \\\cline{2-11}
 &
  ID MMC &
 
  CIFAR-100 &
  80.54&
  81.85 &
  80.64 &
 81.78 &
  \textbf{ 82.21} &
  80.77 &
  59.55&
  47.29 
    \\\cline{2-11}
 &
  \multirow{7}{*}{OOD MMC} 
   &
  SVHN &
  61.15 &
  57.91 &
  39.54 &
  51.02 &
  44.34 &
  32.50 &
  3.96&
  \underline{ 2.26}
   \\
 &
   &
  CIFAR-10 &
  52.13 &
  55.23 &
  54.30 &
  55.58 &
  52.82 &
  49.10 &
  15.56&
  \underline{ 8.49}
   \\
 &
   &
  LSUN\_CR &
  53.19 &
  51.41 &
  54.22 &
  52.34 &
  51.33 &
  48.01 &
  3.10&
  \underline{ 1.60} 
   \\
 &
   &
  Noise &
  61.40 &
  57.89 &
  19.27&
  57.67 &
  19.96 &
 14.92 &
 10.94&
  \underline{7.84} 
   \\
 &
   &
  Uniform &
  59.62 &
 34.05 &
  \textbf{1.00} &
  \textbf{1.00} &
  1.77 &
  \textbf{1.00} &
   2.03&
  \underline{1.00} 
  \\
 &
   &
  Adv. Noise &
  100.00 &
  98.50 &
  1.30 &
  \textbf{1.00} &
  6.50 &
  \textbf{1.00} &
  100.00&
  \underline{1.00} 
   \\
 &
   &
  Adv. Samples &
  99.90 &
  99.90 &
  86.30 &
  - &
  12.90 &
  \textbf{4.30} &
  - &
  - \\ \hline
\multirow{9}{*}{{\rotatebox{90}{SVHN}}} &
  TE  &
  SVHN &
  2.89 &
  2.88 &
  3.05 &
  3.07 &
  3.02 &
  \textbf{ 2.84} &
   4.05&
  3.05 
  \\ \cline{2-11}
 &ID MMC 
   &
  SVHN &
  98.47 &
  98.58 &
  98.52 &
   98.62 &
 98.45 &
  \textbf{98.69 }&
  96.93&
  98.07 
    \\\cline{2-11}
 &
 \multirow{7}{*}{OOD MMC }
   &
  CIFAR-10 &
  71.94 &
  71.70 &
  69.28 &
  68.40 &
  61.09 &
  50.78 &
\underline{10.14} &
  \underline{10.14} \\
 &
   &
  CIFAR-100 &
  71.76 &
  71.04 &
  68.78 &
  68.63 &
  54.09 &
  53.46 &
  \underline{ 10.16}&
  10.20 
  \\
 &
   &
  LSUN\_CR &
  71.27 &
  71.06 &
  62.18 &
  65.78 &
  36.45 &
  29.98 &
  10.14&
  \underline{10.09 }
   \\
 &
   &
  Noise &
  72.00 &
  68.87 &
  39.89 &
  63.43 &
  35.58 &
  \textbf{33.53} &
  35.57&
  48.81 
   \\
 &
   &
  Uniform &
  67.80 &
  40.06 &
  \textbf{10.00} &
  \textbf{10.00} &
  10.34 &
  \textbf{10.00} &
  10.10&
  \underline{10.00} 
   \\
 &
   &
  Adv. Noise &
  100.00 &
  94.60 &
  10.10 &
  \textbf{10.00} &
  24.30 &
  11.00 &
  100.00&
 \underline{10.00} 
   \\
 &
   &
  Adv. Samples &
  100.00 &
  99.50 &
  36.90 &
  - &
  38.70 &
  \textbf{11.40} &
  - &
  - \\ \hline
  
\multirow{9}{*}{{\rotatebox{90}{MNIST}}} &
  TE  &
  MNIST &
  0.51 &
  0.50 &
  0.50 &
  0.49 &
 \textbf{ 0.47} &
  0.51 &
  0.75 &
  0.51 
  \\ \cline{2-11}
 &ID MMC 
   &
  MNIST &
 99.18 &
  99.16 &
  99.15 &
  99.16 &
  98.99 &
  \textbf{99.34} &
   99.27&
 99.16 
  \\\cline{2-11}
 &\multirow{7}{*}{OOD MMC}
   &
  FMNIST &
  66.31 &
  52.88 &
  28.58 &
  63.93 &
  35.32 &
  \textbf{20.98} &
  34.38&
  25.99 
   \\
 &
   &
  EMNIST &
  81.95 &
  81.81 &
  77.92 &
  83.01 &
  69.54 &
  \textbf{47.78} &
  88.00&
  77.74 
   \\
 &
   &
  GrCIFAR-10 &
  46.41 &
  19.10 &
  10.10 &
  10.02 &
  10.43 &
  \textbf{10.00} &
   11.50&
  \underline{10.00} 
  \\
 &
   &
  Noise &
  12.70 &
  12.09 &
  10.36 &
  10.59 &
  10.51 &
  \textbf{10.00} &
  10.22&
  10.34 
  \\
 &
   &
  Uniform &
  97.33 &
  10.01 &
  \textbf{10.00} &
  \textbf{10.00} &
  10.01 &
  10.40 &
  10.01&
  \underline{10.00} 
   \\
 &
   &
  Adv. Noise &
  100.00 &
  14.70 &
  16.20 &
  \textbf{10.00} &
  12.50 &
  \textbf{10.00} &
  100.00&
  \underline{ 10.00} 
   \\
 &
   &
  Adv. Samples &
  99.90 &
  98.20 &
  85.40 &
  - &
  63.80 &
  \textbf{45.20} &
  - &
  - \\ \hline
\multirow{9}{*}{{\rotatebox{90}{FMNIST}}} &
  TE &
  FMNIST &
  4.77 &
  5.01 &
  4.78 &
  4.85 &
 \textbf{  4.56 }&
  4.79 &
   6.12&
  4.96
  \\ \cline{2-11}
 & ID MMC 
   &
  FMNIST &
  98.38 &
  98.24 &
 98.03 &
  98.32 &
   \textbf{98.44} &
  98.35 &
  98.30&
 98.46 
   \\\cline{2-11}
 &\multirow{7}{*}{OOD MMC}
   &
  MNIST &
  71.32 &
  73.44 &
  73.70 &
  71.25 &
  69.67 &
  \textbf{61.47} &
  80.34 &
  70.54 
  \\
 &
   &
  EMNIST &
  65.01 &
  67.34 &
  66.63 &
  68.68 &
  62.97 &
  59.13 &
  36.66&
\underline{ 31.62} 
   \\
 &
   &
  GrCIFAR-10 &
  86.17 &
  69.69 &
  72.90 &
  56.33 &
  66.88 &
  63.24 &
   10.22&
 \underline{10.09} 
  \\
 &
   &
  Noise &
  67.72 &
  57.40 &
  16.75 &
  56.84 &
  14.71 &
  13.03 &
  10.45&
  \underline{10.25} 
   \\
 &
   &
  Uniform &
  77.70 &
  60.08 &
  \textbf{10.00} &
  20.00&
  \textbf{10.00} &
  10.06 &
  73.16&
\underline{10.00} 
   \\
 &
   &
  Adv. Noise &
  100.00 &
  22.30 &
  16.78 &
  \textbf{10.00} &
  14.99 &
  10.18 &
  100.00&
  \underline{ 10.00} 
   \\
 &
   &
  Adv. Samples &
  100.00 &
  99.67 &
  90.56 &
  - &
  70.14 &
  \textbf{59.43} &
  - &
  - \\ \ourline{1pt}
\end{tabular}

}
\caption{We train eight models  on five  datasets and evaluate them on the original dataset and OOD samples, including other datasets, Noise, Uniform, Adversarial (Adv.) Noise and Adversarial Samples. We report the test error (TE) $\downarrow$ of all models, show  mean maximum confidence (MMC)   on in-  and out-of-distribution  samples (e.g.,
ID MMC $\uparrow$ and OOD MMC $\downarrow$).
All values are in percent (\%).
} 
\vspace{-3mm}
\label{exp:MMC}
\end{table*}

\firstpara{Datasets}  Various datasets are used: CIFAR-10, CIFAR-100~\cite{krizhevsky-2009-cifar}, GrCIFAR-10 (gray scale CIFAR-10),   SVHN~\cite{netzer-2011-svhn}, LSUN\_CR (the classroom subset of LSUN~\cite{yu2015lsun}), MNIST, FMNIST~\cite{xiao2017fashion}, EMNIST~\cite{cohen2017emnist}, Noise (i.e., randomly permuting pixels of images from the training set as in~\cite{Meinke-2020-CCU}), Uniform (i.e., uniform noise over the $[0, 1]^d$ box as in~\cite{Meinke-2020-CCU}), 
Adversarial Noise and Adversarial Sample following the experimental setting as~\cite{Meinke-2020-CCU}. 
Adversarial Noise is generated by actively searching for images which yield higher prediction confidences in a neighbourhood of noise images,  while Adversarial Sample is generated in a neighbourhood of in-distribution images but are off the data manifold following~\cite{Hein-2019-WhyOutOfDistribution}. 
For OE and CCUd, we adopt  80 Million Tiny Images~\cite{Torralba-2008-80Million}  with all examples that appear in CIFAR-10 and CIFAR-100 removed as the auxiliary dataset as in~\cite{Meinke-2020-CCU}. 

\firstpara{Methods} Eight methods are evaluated and compared: Baseline, CEDA~\cite{Hein-2019-WhyOutOfDistribution}, ACET~\cite{Hein-2019-WhyOutOfDistribution}, OE~\cite{Hendrycks-2018-AnomalyDetection}, 
two variants of CCU~\cite{Meinke-2020-CCU} (CCUs that adopts noise and CCUd that adopts an auxiliary dataset as in~\cite{Hendrycks-2018-AnomalyDetection}), Ours and Ours++. 
Particularly, 
Ours++ is an enhanced version of Ours,  with selecting the worst cases, i.e., have the largest prediction confidences, in a neighborhood of \oursshort{} similar as in ACET~\cite{Hein-2019-WhyOutOfDistribution}.

\noindent{\bf{Setup}}.  
We train LeNet on MNIST and FMNIST while ResNet-18 for CIFAR-10, CIFAR-100 and SVHN, and then evaluate them on the corresponding test set to report the test error (TE), and on both in- and out-of-distribution datasets to  report mean maximal confidence (MMC) following~\cite{Meinke-2020-CCU}.

\firstpara{Comparisons with the State-of-the-art Methods}
The results in Tab.~\ref{exp:MMC} show that Ours and Ours++ perform the best in most cases without considering  OE and CCUd on  CIFAR-10, CIFAR-100 and SVHN. 
Since training with suppressing the predictions on the large 80 Million Tiny Images~\cite{Torralba-2008-80Million}, that has similar image style as the OOD datasets (e.g., CIFAR-10, CIFAR-100 and SVHN),  OE and CCUd obtain the lowest MMCs. 
However, it could be seen that the auxiliary dataset brings a detrimental impact on the predictions on in-distribution samples, 
e.g., 33\% lower prediction confidences on CIFAR-100 for CCUd, and thus leads to worse  classification performance, e.g., 3.2\% larger test error. Besides, for datasets that have large differences with the auxiliary dataset,  OE and CCUd are comparable with and even worse  than Ours and Ours++, e.g., on FMNIST and MNIST.

Particularly,  ACET performs  better than CEDA, validating the usefulness of  the strategy that searches  harder examples in a neighbourhood of the original ones.
We would like to point out that our method is comparable  with ACET even without picking harder examples,  indicating that \ourssmall{} are more effective than random noises.
By applying the same strategy to Ours, we could see  significant drops in MMC values. 
For  Adversarial Noise and Adversarial Sample, we could  see that CEDA and OE fail in most  cases, 
ACET could handle part of samples, 
while Ours++, CCUs and CCUd perform the best.
Overall, \oursshort{} are effective in alleviating the OOD overconfidence issue.

\begin{figure}[t!]
\centering
  \includegraphics[width=0.99\linewidth]{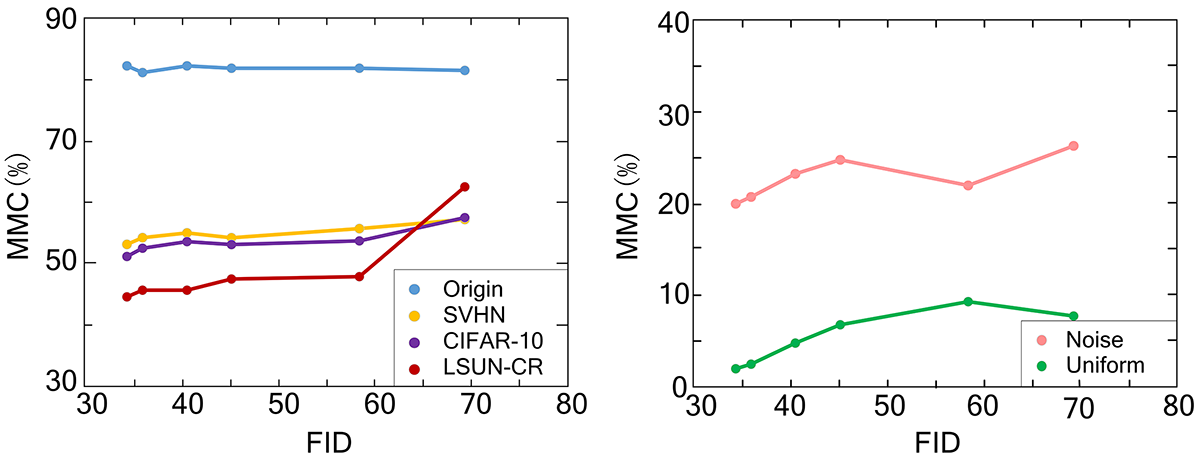}
  \vspace{-1mm}
  \caption{The distribution distances between \ourssmall{} and the training data measured by  FID versus the MMCs on the original dataset (Origin) and various OOD datasets, Noise and Uniform made by ResNet-18 trained on CIFAR-100.}
  \vspace{-2.5mm}
  \label{fig:fid}
\end{figure}

\firstpara{Distribution Distance vs. MMC}
We investigate how distribution distance between \oursshort{} and the in-distribution data would affect the  benefits brought by suppressing \oursshort{} on alleviating the OOD overconfidence issue.
Since we apply a relatively high weight $1.0$ to the Chamfer distance loss, while a low weight $1.0\mathrm{e}{-5}$ to Wasserstein distance loss, the distribution transforming could be carried on gradually during the training process.
To facilitate  fair comparisons, we choose the models of Chamfer GAN saved at different epochs during the training stage on  CIFAR-100,  including models of 
200$th$, 400$th$, 800$th$,  1200$th$, 1600$th$ and 1800$th$, and then train ResNet-18 models with suppressing predictions on the \oursshort{} outputted by the above six Chamfer GANs respectively.
Fig.~\ref{fig:fid} shows that MMCs are mostly positively relevant to the FID scores, and the correlation is stronger on OOD datasets than  on Noise and Uniform,
validating that smaller distribution distance to the in-distribution samples  is critical for effective OOD examples.
We also report the MMCs on in-distribution samples of the six different ResNet-18 models in Fig.~\ref{fig:fid}, and could see the MMCs mostly remain unchanged.

Indeed, OE have also mentioned the influence of distribution distance~\cite{Hendrycks-2018-AnomalyDetection}. 
However,  since the difference between different auxiliary datasets could have many different factors, e.g., RGB values, local textures, it is thus not suitable to conclude which factor affects the result. 
Differently, we transform the distribution of seed examples with Chamfer distance loss to maintain the low-level pixel statistics, such that could rule out the influence of many other factors.

\begin{figure}[t!]
\centering
  \includegraphics[width=0.78\linewidth]{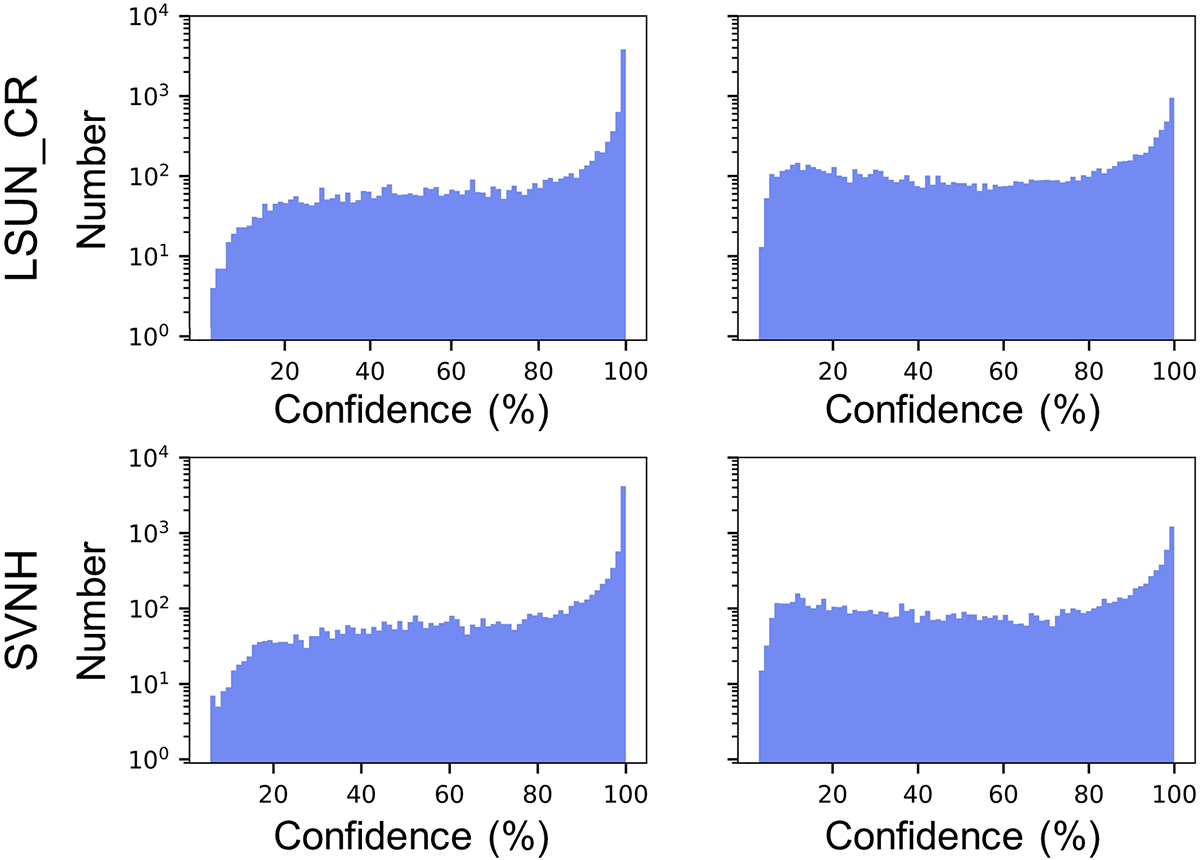}
  \vspace{-1mm}
  \caption{Histograms (logarithmic scale) of maximum confidence values of ResNet-18 trained   for CIFAR-100 on various  datasets.}
  \label{fig:histShow}
\end{figure}

\firstpara{Visualization of Maximum Confidence} 
We visualize the maximum confidences predicted on the images in  LSUN\_CR and SVHN by ResNet-18 trained on  CIFAR-100 using  logarithmic histograms in Fig.~\ref{fig:histShow}.
It could be seen that, by adopting our method, the confidence distributions made by Baseline are pulled to the left, with
the number of samples with high confidence largely reduced.

\firstpara{In-distribution Confidence Calibration} 
We report expected calibration errors (ECEs)~\cite{Guo-2017-OnCalibration} of ResNet-18 trained on CIFAR-10/100 and test on the corresponding test set in Tab.~\ref{exp:ece}. It could be seen  that ECEs are  reduced after applying our method, while temperature tuning is more effective in calibrating prediction confidence on in-distribution data.

\begin{table}[]
\centering
\scalebox{0.9}{ 
\begin{tabular}{ccc|cc}
\ourline{1pt}
\multirow{2}{*}{} & \multicolumn{2}{c|}{CIFAR-10} & \multicolumn{2}{c}{CIFAR-100} \\ \cline{2-5}
                  & Before TT      & After TT     & Before TT      & After TT      \\ \hline
w/o Ours          & 0.033          & 0.031        & 0.081          & 0.073         \\
w/ Ours           & 0.008          & 0.006        & 0.059          & 0.042         \\ \ourline{1pt} 
\end{tabular}
}
\caption{ Expected calibration errors (ECEs) $\downarrow$ on CIFAR-10 and CIFAR-100 w/ and w/o Ours and temperature tuning (TT). }
\label{exp:ece}
\end{table}

\subsection{Applications of \oursshort{}}

\subsubsection{Improving OOD Detection}
\label{subsubsec:OOD_detect}

\firstpara{Evaluation Strategy  and Metrics}
We follow the evaluation strategy as in~\cite{Hendrycks-2017-Detecting-Out-of-Distribution} and use the three common metrics:   the false positive rate  of OOD examples when true positive rate of in-distribution examples is at 95\% ({\bf{FPR95}}),
 the area under the receiver operating characteristic curve ({\bf{AUROC}}), and 
 the area under the precision-recall curve ({\bf{AUPR}}).

\begin{table}[t!]
\centering
\scalebox{0.8}{
\begin{tabular}{c|l|ccc}
\ourline{1pt}
                          &         & {FPR95 $\downarrow$} & {AUROC $\uparrow$} & {AUPR $\uparrow$} \\ \hline
\multirow{4}{*}{CIFAR-10}  & OE      & 8.53          & 98.30          & 99.63          \\
                          & OE+\oursshort{} &\textbf{8.01}            & \textbf{99.07}           & \textbf{99.79}         \\\cline{2-5}
                          & ES      & 3.32           & 98.92          & 99.75         \\
                          & ES+\oursshort{} &\textbf{3.24}           & \textbf{99.01}          & \textbf{99.78}         \\ \hline
\multirow{4}{*}{CIFAR-100} & OE      & 58.10          & 85.19          & 96.40         \\
                          & OE+\oursshort{} & \textbf{56.54}          & \textbf{87.96}          & \textbf{97.38}         \\\cline{2-5}
                          & ES      & 47.55          & 88.46          & 97.12          \\
                          & ES+\oursshort{} & \textbf{45.89}          & \textbf{89.03}          & \textbf{97.95}         \\  \ourline{1pt} 
\end{tabular}
}
\caption{The improvement brought by \oursshort{} on OE~\cite{Hendrycks-2018-AnomalyDetection} and Energy Score (ES)~\cite{Liu-2020-energyOOD}. All values are in percent (\%).}
\vspace{-3mm}
\label{tab:detection_OOD_2}
\end{table}

\firstpara{Methodology}
We  replace the original DNN classifier with the one trained with  suppressing  predictions on \oursshort{}.

\firstpara{Improving OOD detectors}
We first evaluate the improvement on OE~\cite{Hendrycks-2018-AnomalyDetection} and Energy Score~\cite{Liu-2020-energyOOD} that use 80 Million Tiny Images~\cite{Torralba-2008-80Million} as the auxiliary dataset. Specifically,  we train WRN-40-2~\cite{Zagoruyko-2016-wideresnet} on CIFAR-10 and CIFAR-100~\cite{krizhevsky-2009-cifar}, and then test it on six datasets: Textures~\cite{Cimpoi-2014-Textures}, SVHN, Places365~\cite{zhou2017places}, LSUN-Crop~\cite{yu2015lsun}, LSUN-Resize~\cite{yu2015lsun}, and iSUN~\cite{Xu-2015-iSUN} following~\cite{Liu-2020-energyOOD}.  
The averaged results  in Tab.~\ref{tab:detection_OOD_2} show that the performance on all three metrics are improved.
We also evaluate the improvement on  ODIN~\cite{Liang-2018-EnhancingOut-of-distribution} and Mahalanobis distance (Maha)~\cite{Lee-2018-Mahalanobis} that do not  require auxiliary datasets.
The results in Tab.~\ref{tab:detection_OOD_1} show that both ODIN and Maha are improved.
Overall, \oursshort{} could be adopted for improving OOD detectors.

\begin{table}[]
\centering
\center
\scalebox{0.78}{
\begin{tabular}{cc|cc|cc}
\ourline{1pt}
\multicolumn{2}{c|}{\textit{\textbf{}}}                    & ODIN                        & ODIN+\oursshort{} & Maha                         & Maha+\oursshort{} \\ \hline
                & CIFAR-10                & { 95.91} & \textbf{96.90}     & { 97.10} & \textbf{97.34}     \\
                & CIFAR-100                & { 94.82} & \textbf{97.12}    & 96.70                        & \textbf{97.08}     \\
                & LSUN\_CR                & { 96.52} & \textbf{96.96}     & { 97.22} & \textbf{97.97}    \\
                & Noise                   & 82.74 & \textbf{83.01}     & \textbf{ 98.00} & 97.99     \\
\multirow{-5}{*}{\rotatebox{90}{SVHN}}     & Uniform  & 97.90 & \textbf{97.94} & { 97.81} & \textbf{98.01} \\ \hline
                & SVHN                    & { 81.35} & \textbf{84.91}     & { 77.52}  & \textbf{79.63}     \\
                & CIFAR-10                 & { 79.50} & \textbf{83.48}     & { 59.94}  & \textbf{64.74}     \\
                & LSUN\_CR                & { 81.41} & \textbf{82.10}    & { 79.73}  & \textbf{82.99}     \\
                & Noise                   & { 76.84} & \textbf{76.92}     & { 90.61}  & \textbf{90.98}     \\
\multirow{-5}{*}{\rotatebox{90}{CIFAR-100}} & Uniform  & { 93.56} & \textbf{94.87} & { 94.37}  & \textbf{95.90} \\ 
\ourline{1pt}
\end{tabular}

}
\caption{The improvement   brought by \oursshort{} on ODIN~\cite{Liang-2018-EnhancingOut-of-distribution} and Mahalanobis distance (Maha)~\cite{Lee-2018-Mahalanobis} in the AUROC metric (\%) $\uparrow$.}
\vspace{-1mm}
\label{tab:detection_OOD_1}
\end{table}

\begin{figure}[t!]
\centering
  \includegraphics[width=0.97\linewidth]{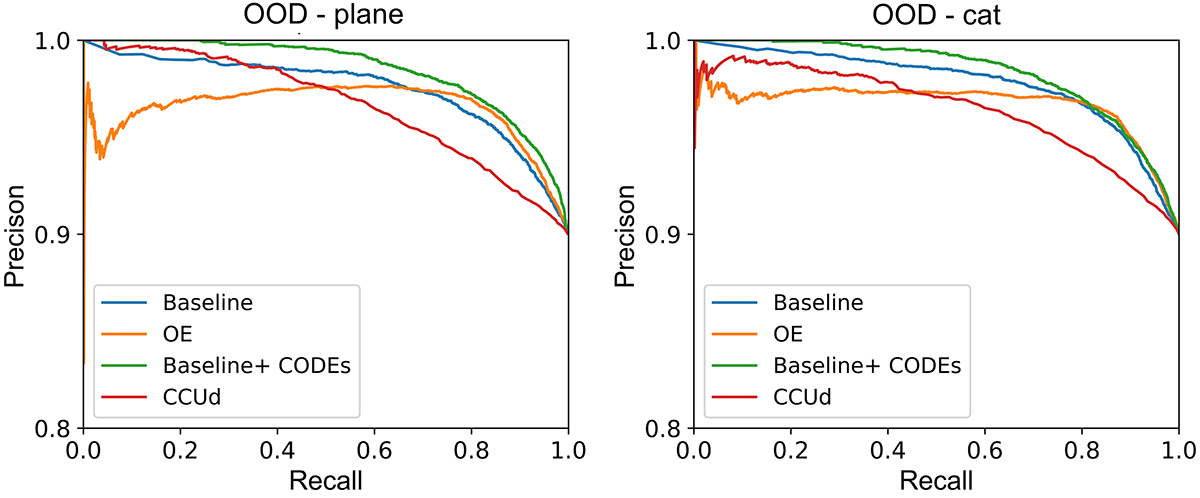}
  \vspace{-1mm}
  \caption{PR curves of four methods for the semantic OOD detection task~\cite{Ahmed-2020-DetectSemanticOOD} on CIFAR-10 with holding out one class as OOD.}
  \vspace{-1mm}
  \label{fig:pr}
\end{figure}

\begin{table}[t!]
\center
\scalebox{0.85}{

\begin{tabular}{cc|cc}
\ourline{1pt}
                          &           & {Baseline} & {Baseline+\oursshort{}}  \\ \hline
\multirow{2}{*}{CINIC-10} & ResNet-32 & 73.82             & \textbf{74.77} \\
                          & ResNet-56 & 74.09             & \textbf{75.38} \\ \hline
\multirow{2}{*}{ImageNet} & ResNet-18 & 69.76             & \textbf{71.06} \\
                          & ResNet-50 & 76.15             & \textbf{77.12} \\ \ourline{1pt} 
\end{tabular}

}
\caption{Top-1 Acc (\%) $\uparrow$ on CINIC-10 and ImageNet.}
\vspace{-3mm}
\label{tab:rec_imagenet}
\end{table}

\firstpara{Detecting Semantic OOD Examples}
We  evaluate the situation where in-distribution samples are not only significantly outnumber OOD ones, but also have significant semantic shifts following~\cite{Ahmed-2020-DetectSemanticOOD}.
Specifically, we train two classifiers for CIFAR-10 with holding out one class every time (e.g., plane, cat), and then score the ability to detect the held out class as OOD samples.
The precision-recall curves are presented in Fig.~\ref{fig:pr}. 
It could be seen that OE~\cite{Hendrycks-2018-AnomalyDetection} and CCUd~\cite{Meinke-2020-CCU} that adopt auxiliary datasets hurt the performance of semantic OOD detection since  predictions on  in-distribution  samples  are suppressed as reported in Tab.~\ref{exp:MMC}, while utilizing \oursshort{} is beneficial to  it.

\subsubsection{Improving Classification} 
We demonstrate \oursshort{} could improve  classification by evaluating ResNets on  CINIC-10~\cite{darlow2018cinic} and ImageNet~\cite{deng2009imagenet}.
Particularly, a separate batch norm for \oursshort{} is adopted following~\cite{Xie-2020-adversarial-improve-recognition}, which is critical for consistent improvement.
The results in Tab.~\ref{tab:rec_imagenet} show that \ourssmall{}  bring 1-2 percent improvement on the top-1 accuracy.
The reason is probably that  \oursshort{} are sampled in  between the decision boundaries of multiple categories since are spliced with   patches from different-category images,  and thus could help to prevent confusion between multiple different categories.

\begin{table}[t!]
\center
\scalebox{0.83}{ 
\begin{tabular}{c|ccc}
\ourline{1pt}
\textbf{}                     & CIFAR-100      & CIFAR-10       & SVHN           \\ \hline
\multicolumn{1}{l|}{Baseline} & 78.99          & 57.50          & 70.95          \\
w/o Chamfer GAN               & 50.23          & 58.08          & 59.67          \\
w/ Chamfer GAN                & \textbf{43.64} & \textbf{54.44} & \textbf{51.62} \\ \ourline{1pt} 
\end{tabular}
}
\caption{Ablation study on Chamfer GAN  with averaged MMC  (\%) $\downarrow$ tested on  OOD datasets listed in Tab.~\ref{exp:MMC}.}
\label{tab:ablation}
\vspace{-2mm}
\end{table}

\begin{table}[t!]
\center
\scalebox{0.85}{ 
\begin{tabular}{l|llll}
\ourline{1pt}
\multicolumn{1}{c|}{{$k$}} & \multicolumn{1}{c}{2} & \multicolumn{1}{c}{4} &  \multicolumn{1}{c}{6}     & \multicolumn{1}{c}{8} \\ \hline
MMC                             & \textbf{43.64}        & 47.89                 & 46.32 & 48.39                 \\ 
\ourline{1pt}
\end{tabular}}
\caption{Averaged MMC (\%) $\downarrow$ of ResNet-18 (trained on CIFAR-100) tested on OOD datasets listed in Tab.~\ref{exp:MMC}.}
\label{tab:ablation_piece}
\vspace{-3mm}
\end{table}

\subsection{Ablation Studies}

\firstpara{Setup}
We train different ResNet-18 models  with suppressing predictions on  \oursshort{} generated with/without Chamfer GAN, and with different   $k$s in the slicing\&splicing operations, and then test them on 
 OOD datasets listed in Tab.~\ref{exp:MMC} 
to report averaged MMC results in Tab.~\ref{tab:ablation} and Tab.~\ref{tab:ablation_piece}.

\firstpara{Chamfer GAN}
It could be seen that ResNet-18 models trained with ablating Chamfer GAN  still improve the performance, since the distribution of seed examples are originally close to the ID distribution with the novel slicing\&splicing operation as reported in Tab.~\ref{exp:MMC}. However, the performance is much worse than that with Chamfer GAN, validating the importance of  distribution transformation.

\firstpara{{Piece $k$}}
 It could be seen that $k=2$ brings the best performance in Tab.~\ref{tab:ablation_piece}, since  larger $k$ may bring too more flexibility for Chamfer GAN to maintain the pixel-level statistics.

\section{Conclusion}

This paper has proposed  \oursshort{},  a kind of  effective  \ourscommon{}  that could be utilized to alleviate the OOD overconfidence issue inherently by suppressing predictions on them.
The key idea of generating CODEs is to restrict the distribution of 
spliced OOD examples generated from training data,  to be close to that of in-distribution samples by Chamfer GAN.
Extensive experiments validate the effectiveness of \oursshort{} and their usefulness in improving OOD detection and classification.
We hope \oursshort{}  inspire more research on alleviating the OOD overconfidence issue.

\vspace{1mm}
\noindent{\textbf{Acknowledgements:}}
\small{
This work was supported  in part by
the Guangdong Basic and Applied Basic Research Foundation (2020A1515110997), 
the 
NSFC
(U20B2046, 61902082, 62072126),
the Science and Technology Program of Guangzhou  (202002030263, 202102010419), 
the Guangdong Higher Education Innovation Group (2020KCXTD007)
and the Talent Cultivation Project of Guangzhou University (XJ2021001901). 
    }

{\small
\bibliographystyle{ieee_fullname}
\bibliography{egbib}
}

\end{document}